\documentclass[sigconf,screen]{acmart}

\usepackage{graphicx} 
\usepackage{overpic} 
\usepackage{amsmath}
\usepackage{algorithm}
\usepackage{algorithmic}
\usepackage{amsfonts}
\usepackage{multirow} 
\usepackage{longtable} 
\usepackage{array} 
\usepackage{makecell}
\usepackage{threeparttable} 
\usepackage{booktabs}

\def\ourmodel{UQFormer}
\usepackage{newfloat}
\usepackage{listings}
\usepackage{graphicx}

\definecolor{myblue}{RGB}{0,102,204}

\usepackage{colortbl} 
\AtBeginDocument{%
  }

\copyrightyear{2023}
\acmYear{2023}
\setcopyright{acmlicensed}
\acmConference[MM '23] {Proceedings of the 31st ACM International Conference on Multimedia}{October 29--November 3, 2023}{Ottawa, ON, Canada}
\acmBooktitle{Proceedings of the 31st ACM International Conference on Multimedia (MM '23), October 29--November 3, 2023, Ottawa, ON, Canada}
\acmPrice{15.00}
\acmISBN{979-8-4007-0108-5/23/10}
\acmDOI{10.1145/XXXXXXX.XXXXXXX}

\renewcommand\footnotetextcopyrightpermission[1]{} 

\begin{document}

\title{A Unified Query-based Paradigm for Camouflaged Instance Segmentation}

\author{Bo Dong}
\authornote{Both authors contributed equally to this research.}
\affiliation{College of Biomedical Engineering and Instrumental Science, 
\institution{Zhejiang University}
  \city{Hangzhou}
  \state{}
  \country{China}
}
\email{dongbo811@zju.edu.com}

\author{Jialun Pei}
\authornotemark[1]
\affiliation{Department of Computer Science and Engineering, 
\institution{The Chinese University of Hong Kong}
  \city{Hong Kong SAR}
  \state{}
  \country{China}
}
\email{jialunpei@cuhk.edu.hk}

\author{Rongrong Gao}
\affiliation{Department of Computer Science and Engineering, 
\institution{The Hong Kong University of Science and Technology}
  \city{Hong Kong SAR}
  \state{}
  \country{China}
}
\email{rgaoaf@connect.ust.hk}

\author{Tian-Zhu Xiang}
\authornote{Corresponding author. Tian-Zhu Xiang (tianzhu.xiang19@gmail.com)}
\affiliation{ 
\institution{G42}
  \city{Abu Dhabi}
  \state{}
  \country{United Arab Emirates}
}
\email{tianzhu.xiang19@gmail.com}

\author{Shuo Wang}
\affiliation{
\institution{ETH Zurich}
  \city{Zurich}
  \state{}
  \country{Switzerland}
}
\email{shawnwang.tech@gmail.com}

\author{Huan Xiong}
\affiliation{
\institution{MBZUAI}
  \city{Abu Dhabi}
  \state{}
  \country{United Arab Emirates}
}
\email{huan.xiong.math@gmail.com}

\renewcommand{\shortauthors}{Bo Dong, et al.}

\begin{abstract}
Due to the high similarity between camouflaged instances and the background, the recently proposed camouflaged instance segmentation (CIS) faces challenges in accurate localization and instance segmentation. 
To this end, inspired by query-based transformers, we propose a unified query-based multi-task learning framework for camouflaged instance segmentation, termed \textbf{UQFormer}, which builds a set of mask queries and a set of boundary queries to learn a shared composed query representation and efficiently integrates global camouflaged object region and boundary cues, for simultaneous instance segmentation and instance boundary detection in camouflaged scenarios.  
Specifically, we design a \emph{composed query learning paradigm} that learns a shared representation to capture object region and boundary features by the cross-attention interaction of mask queries and boundary queries in the designed multi-scale unified learning transformer decoder. 
Then, we present a \emph{transformer-based multi-task learning framework} for simultaneous camouflaged instance segmentation and camouflaged instance boundary detection based on the learned composed query representation, which also forces the model to learn a strong instance-level query representation. Notably, our model views the instance segmentation as a query-based direct set prediction problem, without other post-processing such as non-maximal suppression. 
Compared with 14 state-of-the-art approaches, our UQFormer significantly improves the performance of camouflaged instance segmentation. Our code will be available at \textcolor{myblue}{\url{https://github.com/dongbo811/UQFormer}}.
\end{abstract}



\begin{teaserfigure}
  \vspace{2mm}
\end{teaserfigure}

\maketitle

\section{Introduction}
Camouflage is a common adaptation that many prey species have adopted in order to reduce the likelihood of being detected or recognized by would-be predators~\cite{skelhorn2016cognition}. The high similarity between camouflaged objects and their background makes camouflaged object detection (COD)~\cite{fan2021concealed,chen2023diffusion} far more challenging than generic object detection~\cite{huang22scribble, Ge23tcnet, Zheng23memory}. In recent years, COD has attracted increasing attention from the computer vision community due to its wide range of applications in our real life, such as wildlife conservation~\cite{nafus2015hiding}, medical image segmentation~\cite{fan2020pra,li2022trichomonas,li2021mvdi25k}, and industrial defect detection~\cite{yang2020using}. Although COD is dedicated to identifying all camouflaged objects from the background at the object level~\cite{pang2022zoom, huang2023Feature}, it falls short of meeting instance-level perception and applications such as object counting and ranking.

\begin{figure}[t!]
	\centering
    \small	\begin{overpic}[width=1.\linewidth]{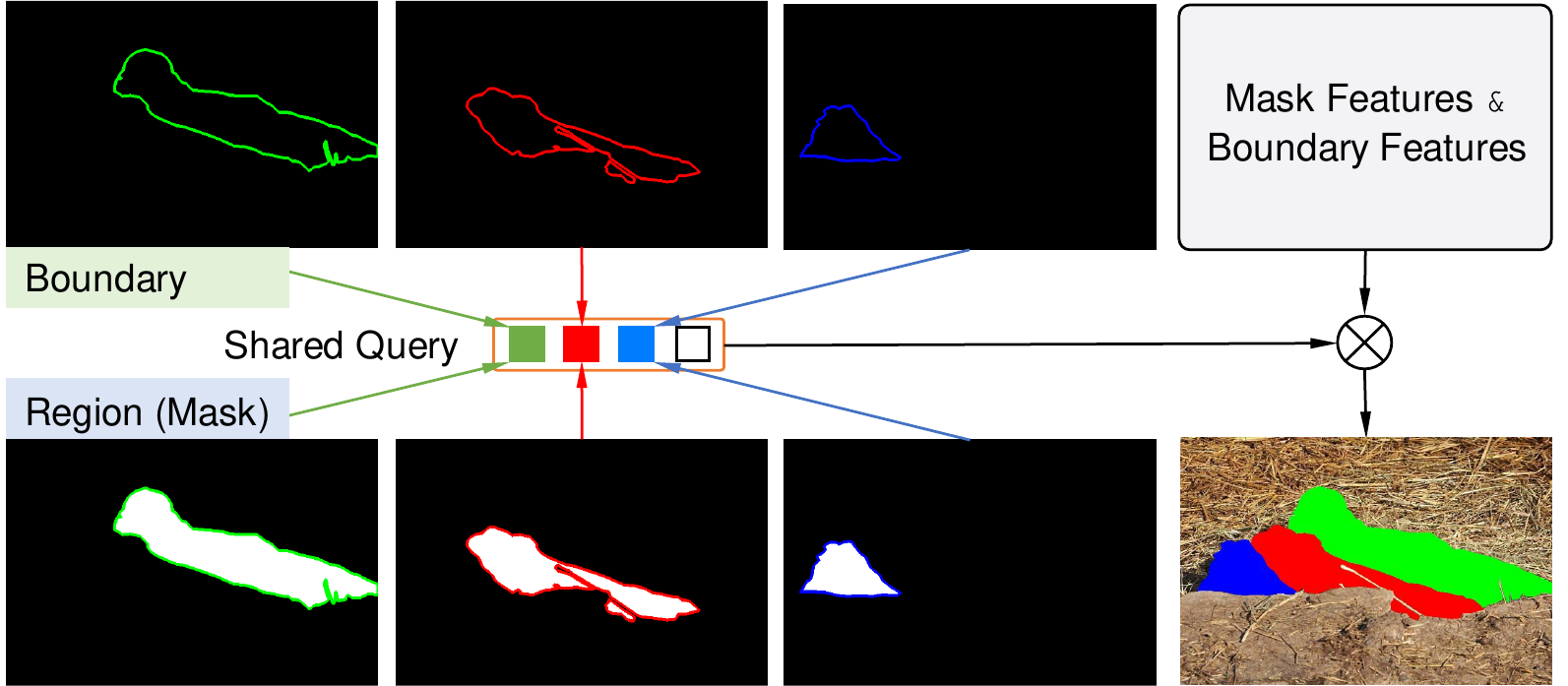}
    \end{overpic}
	\caption{\textbf{Illustration of the proposed~\ourmodel~for camouflaged instance segmentation. \ourmodel~integrates boundary and region (mask) features of camouflages to build strong shared query feature representations for instance-level predictions using the multi-task query-based transformers.} 
    } 
    \label{figure: idea}
\end{figure}

To address this limitation, camouflaged instance segmentation (CIS)~\cite{le2021camouflaged,pei2022osformer} has been introduced to detect and delineate each distinct camouflaged instance in an image at the pixel level. 
From the perspective of generic instance segmentation, current approaches can be broadly divided into three categories, ``detect-then-segment'' strategy (\textit{e.g.}, Mask R-CNN~\cite{he2017mask}), ``label-then-cluster'' strategy (\textit{e.g.}, SSAP~\cite{gao2019ssap}), and direct instance segmentation (\textit{e.g.}, SOLO~\cite{wang2020solo}). 
The first two manners heavily rely on precise bounding boxes or per-pixel embedding learning and clustering processing, which are indirect and step-wise. 
Hence, the direct manner has received more attention lately. 
However, in contrast to generic instance segmentation, CIS is defined as a \textit{de facto} class-agnostic instance segmentation task. It is carried out in more concealed scenarios where the backgrounds are indistinguishable, making CIS far trickier. 
To our knowledge, CIS remains an under-explored issue as little effort is devoted to instance-level partitioning. 
Recently,~\cite{pei2022osformer} proposed the first transformer-based framework (\textit{i.e.}, OSFormer) for CIS, but it still requires hand-designed NMS-like post-processing and provides limited performance.

In the transformer-based architecture~\cite{carion2020detr,zhu2020deformable}, object queries play a critical role in transformer decoders, which are used a) as candidates and the initial input to the decoder and b) to interact with transformer encoder features in the decoder layers for generating output embeddings. 
We argue that a meticulously designed query feature can effectively enhance the aggregation of camouflaged features. 
However, most existing methods adopt random initialization~\cite{carion2020detr}, which leads to slow training convergence and affect the performance of feature learning. 
For segmenting camouflaged instances, OSFormer designs location-guided queries, which utilize learnable local features in different locations for query initialization to improve performance. Nevertheless, it is worthwhile to further investigate \textit{how to design a well-performing query for camouflaged learning.}
Furthermore, recent studies~\cite{zhai2021mutual,sun2022bgnet} demonstrate the critical role of the boundary in COD.
Despite the complexity of camouflaged patterns, in most cases, the object boundary typically contains valuable clues that can distinguish objects from the background. \textit{Thus, it is essential to explore boundary features and incorporate them into transformer models for CIS.}

Inspired by query-based transformers, we propose a unified query-based multi-task learning framework for CIS, termed \textit{UQFormer}. Our model introduces multiple kinds of queries to learn a shared composed query for simultaneous instance segmentation and instance boundary detection.
Similar to DETR~\cite{carion2020detr}, we formulate the instance segmentation as a query-based direct set prediction problem, which avoids the post-processing such as non-maximal suppression~\cite{he2017mask}.
To build powerful object queries, we introduce two kinds of queries, \textit{i.e.}, mask queries and boundary queries, to interact with global features (including object region and boundary features) and then generate a composed query for the instance decoding, by the designed multi-scale unified learning transformer decoder. 
The proposed query learning paradigm efficiently aggregates camouflaged region and boundary features to boost model performance and accelerate training convergence.
More importantly, we construct a transformer-based multi-task learning framework that supports both instance segmentation and instance boundary detection in camouflaged scenarios. These two tasks are associated with a shared composed query representation. 
\ourmodel~forces the shared query to learn a robust instance-level query representation by leveraging the guidance of two highly related tasks.  
This design enables cross-task communication and collaboration between instance segmentation and instance boundary detection, thus encouraging the two tasks to benefit from each other. 
Benefiting from cross-task composition query learning and multi-task joint learning, \ourmodel~can build a powerful query feature to integrate effective clues to object regions and boundaries and model the relationship between each object instance, providing outstanding performance.

Our main contributions can be summarized as follows: 

\begin{itemize}
    \item We propose a composed query learning paradigm, which learns a shared query feature and efficiently aggregates camouflaged region and boundary features from mask query and boundary query through the multi-scale unified learning transformer decoder. 
    
    \item A multi-task learning framework is designed for simultaneous instance segmentation and instance boundary detection, which views instance-level segmentation as a direct set prediction problem.
    
    \item Extensive experiments demonstrate the effectiveness of the proposed UQFormer on the challenging CIS task and the superior performance over 14 competitors. 
   
\end{itemize}

\begin{figure*}[t!]
	\centering
    \small
	\begin{overpic}[width=1.0\linewidth]{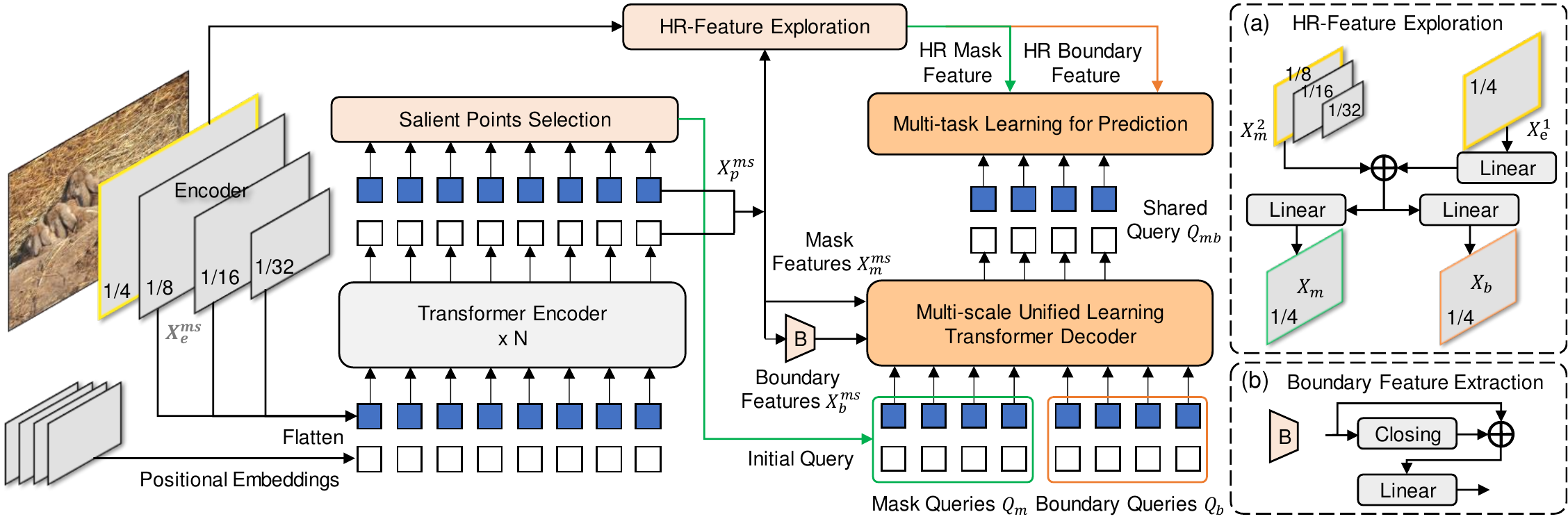}
    \end{overpic}
\caption{\textbf{Overall architecture of our~\ourmodel.} Given an input image, we first adopt the CNN backbone to extract multi-scale features and use the transformer encoder to enhance the features. Based on the enhanced features, we initialize the object queries by a salient points selection method. Then, a multi-scale unified learning transformer decoder is designed to learn the shared composed query by fully exploring and interacting mask and boundary cues. Finally, a transformer-based multi-task learning module is embedded to integrate the composed query feature and high-resolution backbone features for camouflaged instance segmentation and camouflaged instance boundary detection.}
    \label{figure: network}
\end{figure*}

\section{Related Works}

\subsection{Instance Segmentation}

The purpose of instance segmentation is to detect objects in the scene and assign pixel-level binary labels. Early instance segmentation methods are based on two-stage patterns, which segment instances in the detected bounding boxes, such as Mask-RCNN~\cite{he2017mask}, Mask Scoring R-CNN~\cite{huang2019mask}, Cascade R-CNN~\cite{cai2018cascade} and HTC~\cite{chen2019hybrid}. 
However, these two-stage models show relatively slow inference. 
In recent years, one-stage models have achieved comparable results to two-stage models with largely simplified detection pipelines, like YOLACT~\cite{bolya2019yolact}, YOLACT++~\cite{bolya2020yolactplus}, BlendMask~\cite{chen2020blendmask}, and CondInst~\cite{tian2020conditional}. These methods generate instance masks by clustering the per-pixel embeddings into an arbitrary number of instances in an image.  
Different from the previous one-stage methods, some instance segmentation methods are proposed to predict instance masks directly without depending on grouping post-processing, e.g., SOTR~\cite{guo2021sotr}, SOLO~\cite{wang2020solo}, SOLOv2~\cite{wang2020solov2}, QueryInst~\cite{fang2021instances}, SparseInst~\cite{cheng2022sparse},
Mask Transfiner~\cite{ke2022mask}, Mask2Former \cite{cheng2021mask2former}, OSFormer~\cite{pei2022osformer}. For instance, SOLO~\cite{wang2020solo} removes the region proposal network and combines the grid positive and negative sample allocation strategy to achieve instance prediction, by only the classification branch and the mask branch. 
SparseInst~\cite{cheng2022sparse} proposes a set of sparse instance activation maps as new object representations to highlight informative regions of each foreground object. 
Furthermore, Mask2Former~\cite{cheng2021mask2former} introduces learnable queries and learns localized features through constraining cross-attention with predicted mask regions. 
For the CIS task, OSFormer~\cite{pei2022osformer} is the first one-stage transformer model, which adopts a location-sensing transformer with coarse-to-fine fusion to predict camouflaged instances. 
Unlike OSFormer, we handle the CIS task from the perspective of query learning, which fully integrates and interacts object region queries and object boundary queries to strengthen the query feature representation. 

\subsection{Query-Based Models}

DETR~\cite{carion2020detr} first proposes the concept of object queries as a learnable embedding, independent of the content of the current input image. Specific object features are clustered through cross-attention, and relationships between instances are established and optimized through self-attention and the feed-forward network. Many works~\cite{cheng2021mask2former,li2022dn,zhu2020deformable} have demonstrated the effectiveness of queries-based learning in different computer vision tasks.
The queries-based model, Maskformer~\cite{cheng2021per}, effectively solves the problem of semantic segmentation and panoramic segmentation. The SeqFormer~\cite{seqformer} obtains the position cues of each object in independent frames by separating the instance queries into box queries and then aggregates them for better instance representation at the video level. Fashionformer~\cite{xu2022fashionformer} simplifies the traditional multi-head architecture through dual-stream synchronous learning of object queries and attribute queries. The essence is to continuously optimize the queries by interacting with the features of the backbone or encoder, making the resulting queries learn the global semantics of specific objects for better representations. 
In this paper, we introduce a multi-task learning framework that incorporates object region queries and object boundary queries to improve the performance of camouflaged instance segmentation.

\section{Proposed Method}

\subsection{Feature Encoder}
The overall framework of the proposed~\ourmodel~is shown in Figure~\ref{figure: network}. First, we build a feature extractor that contains a CNN backbone (\textit{e.g.}, ResNet-50~\cite{he2016deep}) and a transformer encoder. Different from using single-scale features in DETR~\cite{carion2020detr} and Maskformer~\cite{cheng2021per}, we adopt deformable attention~\cite{zhu2020deformable} to compute the interaction between multi-scale features. 

Specifically, we utilize the CNN network to extract the multi-scale features $X_{e}^{ms}=\{X_{e}^{1}, ..., X_{e}^{i}\}, i \in \{1,2,3,4\}$ from input images $I \in \mathbb{R}^{H\times W\times 3}$. Then the output features ($X_e^2\sim X_e^4$) are concatenated and then fed into the transformer encoder to obtain the multi-scale transformer features $X_{p}^{ms}=\{X_{p}^{2}, ..., X_{p}^{i}\}, i \in \{2,3,4\}$.
For a fair comparison, we set the number of transformer encoder layers to 6.
On this basis, we construct multi-scale mask features $X_{m}^{ms}$ and multi-scale boundary features $X_{b}^{ms}$ for query learning, which can be calculated as:
\begin{equation}
\left\{ \begin{aligned}
&X_{m}^{ms}=(X_{p}^{ms}), \\
&X_{b}^{i}=f_{b}^{lm}(f_{b}^{co}(X_{p}^{i}) + X_{p}^{i}), i \in \{2,3,4\},
\end{aligned} 
\right.
\end{equation}
where $f_{b}^{lm}$ is a linear mapping layer, implemented with the $1 \times 1$ convolution layer. $f_{b}^{co}$ represents the closing operation, which uses a fixed convolution kernel to process each channel. The closing operation can eliminate the noise points inside the camouflaged objects and avoid misleading the learning of boundary queries.

\subsection{Multi-scale Unified Query Learning}

\subsubsection{Query Initialization}
Due to the difference between the tasks of camouflaged instance segmentation and camouflaged instance boundary detection, we construct separate queries for each task rather than directly using the same queries for these two tasks, denoted as $Q_{m}$ and $Q_{b}$. 
Existing works generally regard the query-based mechanism as a clustering strategy. Using appropriate seed points as the initial clustering center can accelerate the convergence of the model, and provides a better clustering performance. 

To generate mask queries, considering the transformer global features have important location information, we directly choose the salient points from the global features to initialize $Q_{m}$, denoted as $Q_{m}^{0}$. 
First, we integrate the multi-scale features $X_{m}^{ms}$ to produce location activation maps. Here we adopt the strategy in Pointrend~\cite{kirillov2020pointrend}. 
Specifically, we select salient points with high confidence instead of uncertain object edge points in Pointrend:
\begin{equation}
Q_{m}^{0} = \texttt{Topk}(\sum f_{m}^{int}(X_{m}^{i})), i \in \{2,3,4\},
\end{equation}
where $f_{m}^{int}$ denotes a convolutional layer to integrate the multi-scale features, and $k$ is equal to the number of queries.

For camouflaged instance boundary detection, we refer to DETR \cite{carion2020detr}, where the boundary queries are initialized randomly and have nothing to do with the extracted transformer features.

\begin{figure}[t!]
	\centering
    \small	\begin{overpic}[width=1.\linewidth]{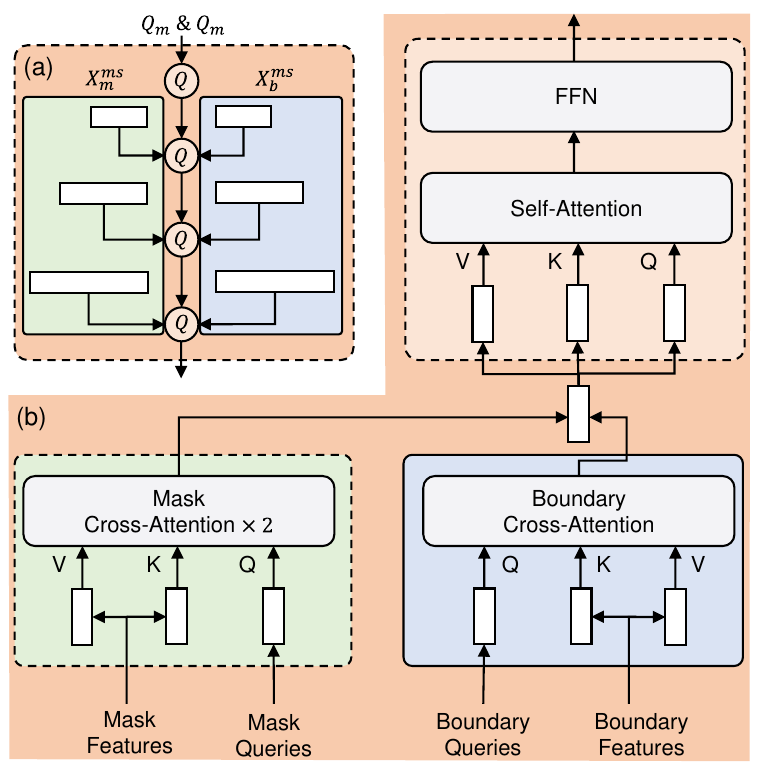}
    \end{overpic}
    \caption{\textbf{The proposed multi-scale unified learning transformer decoder.} (a) shows the overall structure which explores queries in a cascaded manner. (b) details query learning/update at each scale. The designed module integrates the mask cues and boundary cues by cross-attention and explores the relationships between camouflaged instances by self-attention, to learn a composed query with strong feature representation. The learned query effectively integrates camouflaged region and boundary cues to boost feature representation. 
    }
    \label{figure: cross_att}
\end{figure}

\subsubsection{Query Interactive Learning}
To learn a strong shared query representation in query learning, we design multi-scale unified learning transformer decoder, which interacts two kinds of queries, \textit{i.e.}, mask queries and boundary queries, with the transformer encoder features to enhance the feature representation. Figure~\ref{figure: cross_att} shows the unified update strategy of mask queries and boundary queries, respectively. Here we adopt a cascaded manner to optimize the queries by multi-scale features. 

Following DETR~\cite{carion2020detr}, we use the fixed position encoding to encode multi-scale mask features $X_{m}^{ms}=\{X_{m}^{2}, ..., X_{m}^{i}\}$ and multi-scale boundary features $X_{b}^{ms}=\{X_{b}^{2}, ..., X_{b}^{i}\}$ respectively, where $i \in \{2,3,4\}$. 
For each scale (\textit{i.e.}, $i$), the mask queries first explore mask features from $X_{m^{\prime}}^{i}$ through cross-attention:
\begin{equation}
Q_{m^{\prime}}^{i} = \texttt{CA}(Q_{m}^{i}+P_{mq}^{i}, X_{m^{\prime}}^{i}),
\end{equation}
where \texttt{CA}~($\cdot$) denotes the standard cross-attention in transformers. $Q_{m}^{i}$ and $Q_{m^{\prime}}^{i}$ denote the learned mask queries from the last iteration and the current iteration, respectively. $P_{mq}^{i}$ denotes the position encoding at $i$ scale. To obtain the feature $X_{m^{\prime}}^{i}$, a convolution is first applied on the feature $X_{m}^{i}$, which is then added with a level embedding. 
Similarly, the camouflaged instance boundary queries $Q_{b^{\prime}}^{i}$ is calculated as:
\begin{equation}
Q_{b^{\prime}}^{i} = \texttt{CA}(Q_{b}^{i}+P_{bq}^{i}, X_{b^{\prime}}^{i}),
\end{equation}
where $Q_{b}^{i}$ and $Q_{b^{\prime}}^{i}$ denote the learned boundary queries from the last iteration and the current iteration. $P_{bq}^{i}$ denotes the position encoding at $i$ scale. A convolution is applied on the feature $X_{b}^{i}$, which is then added with a level embedding to generate the feature $X_{b^{\prime}}^{i}$. 
The mask cross-attention adopts two layers and the boundary cross-attention adopts one layer at each scale.

Mask queries and boundary queries represent camouflaged instance features from the perspective of region and boundary, respectively. Thus, we combine these two learned queries and then adopt a standard multi-head self-attention (\texttt{MHSA}) and a standard feed-forward network (\texttt{FFN}) to generate a composed query $Q_{mb}^{i}$. It can be described as: 
\begin{equation}
\left\{\begin{aligned}
&Q_{mb^{\prime}}^{i} = Q_{m^{\prime}}^{i} +Q_{b^{\prime}}^{i}, \\
&Q_{mb}^{i} = \texttt{FFN}(\texttt{MHSA}(Q_{mb^{\prime}}^{i})).
\end{aligned} \right.
\end{equation}
The composed query $Q_{mb}^{i}$ will be used as the mask queries and boundary queries in the next stage, respectively. 
With the proposed multi-scale unified learning strategy at each scale, mask queries and boundary queries fully interact with mask features and boundary features. 
The proposed module combines the mask and boundary cues and captures the correlations among camouflaged instances to learn a composed query with enhanced feature representation.

\begin{figure}[t!]
	\centering
    \small	\begin{overpic}[width=1.\linewidth]{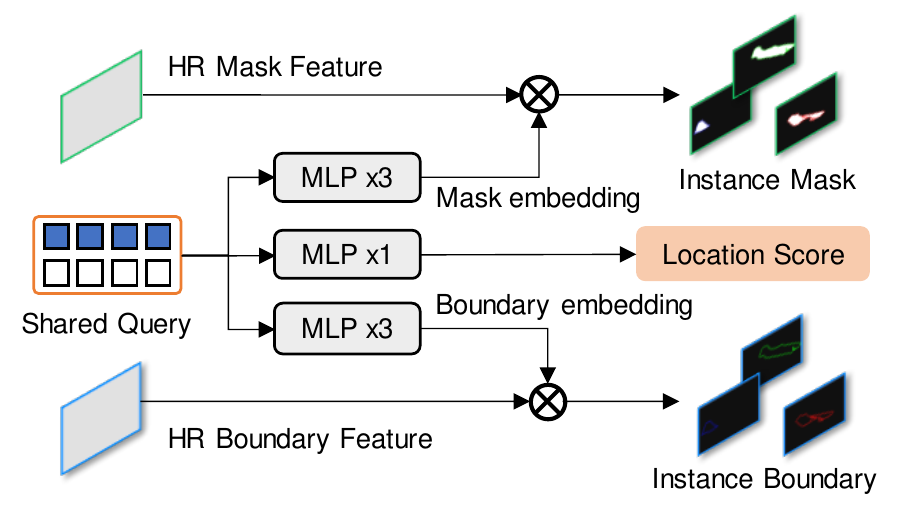}
    \end{overpic}
\caption{Multi-task learning module for instance prediction. The module predicts camouflaged object instances and boundary instances.}
    \label{figure: prediction}
\end{figure}

\begin{table*}[t!]
\small 
\caption{Quantitative comparison of CIS with 14 SOTA methods on COD10K and NC4K. We report the results on ResNet-50 and ResNet-101 as backbones. The best and second-best results are bolded and underlined respectively.}
\centering
\renewcommand{\arraystretch}{1.0}
\setlength\tabcolsep{2.5mm}
\resizebox{1\textwidth}{!}{
\begin{tabular}{l|ccc|ccc|ccc|ccc}
\toprule
Backbone&\multicolumn{6}{c|}{ResNet-50} & \multicolumn{6}{c}{ResNet-101}\\
\hline
Dataset&\multicolumn{3}{c|}{COD10K} &\multicolumn{3}{c|}{NC4K} &\multicolumn{3}{c|}{COD10K} &\multicolumn{3}{c}{NC4K} \\
\cline{1-13}
Metric& $\rm AP$ &$\rm AP_{50}$&$\rm AP_{75}$&  $\rm AP$ &$\rm AP_{50}$&$\rm AP_{75}$& $\rm AP$ &$\rm AP_{50}$&$\rm AP_{75}$&  $\rm AP$ &$\rm AP_{50}$&$\rm AP_{75}$ \\
\midrule
Mask R-CNN~\cite{he2017mask}&25.0&55.5&20.4&27.7&58.6&22.7&28.7&60.1&25.7&36.1&68.9&33.5 \\

MS R-CNN~\cite{huang2019mask}&30.1&57.2&28.7&31.0&58.7&29.4&33.3&61.0&32.9&35.7&63.4&34.7 \\

Cascade R-CNN~\cite{cai2018cascade}&25.3&56.1&21.3&29.5&60.8&24.8&29.5&61.0&25.9&34.6&66.3&31.5 \\

HTC~\cite{chen2019hybrid}&28.1&56.3&25.1&29.8&59.0&26.6 &30.9&61.0&28.7&34.2&64.5&31.6 \\

BlendMask~\cite{chen2020blendmask}&28.2&56.4&25.2&27.7&56.7&24.2 &31.2&60.0&28.9&31.4&61.2&28.8\\

Mask Transfiner~\cite{ke2022mask}&28.7&56.3&26.4&29.4&56.7&27.2 &31.2&60.7&29.8&34.0&63.1&32.6 \\

YOLACT~\cite{bolya2019yolact}&24.3&53.3&19.7&32.1&65.3&27.9 &29.0&60.1&25.3&37.8&70.6&35.6 \\

CondInst~\cite{tian2020conditional}&30.6&63.6&26.1&33.4&67.4&29.4 &34.3&67.9&31.6&38.0&71.1&35.6\\

QueryInst~\cite{fang2021instances}&28.5&60.1&23.1&33.0&66.7&29.4 &32.5&65.1&28.6&38.7&72.1&37.6 \\

SOTR~\cite{guo2021sotr}&27.9&58.7&24.1&29.3&61.0&25.6 &32.0&63.6&29.2&34.3&65.7&32.4 \\

SOLOv2~\cite{wang2020solov2}&32.5&63.2&29.9&34.4&65.9&31.9 &35.2&65.7&33.4&37.8&69.2&36.1 \\

SparseInst~\cite{cheng2022sparse}&32.8&60.5&31.2&34.3&61.3&32.8&36.0&63.2&35.4&38.3&65.9&37.8  \\
Mask2Former~\cite{cheng2021mask2former}&\underline{41.4} &68.5 &\underline{41.6} &\underline{44.6} &71.7 &\underline{45.7} &\underline{44.3} &70.5 &\underline{46.0} &\underline{49.2} &71.6 &\underline{51.4}  \\ 
\hline
OSFormer~\cite{pei2022osformer}  &41.0 &\underline{71.1} &40.8 &42.5 &\underline{72.5} &42.3 &42.0 &\underline{71.3} &42.8 &44.4 &\underline{73.7} &45.1  \\
\rowcolor{gray!20}
\textbf{\ourmodel~(Ours)} &\textbf{45.2}&\textbf{71.6}&\textbf{46.6}&\textbf{47.2}&\textbf{74.2}&\textbf{49.2} &\textbf{45.4}&\textbf{71.8}&\textbf{47.9}&\textbf{50.1}&\textbf{76.8}&\textbf{52.8} \\

\bottomrule
\end{tabular}}
\label{table: all_data}
\end{table*}

\subsection{Multi-task Learning}

\subsubsection{High-Resolution Feature Exploration}
The high-resolution (HR) features are explored to complement detail cues for learned composed query embedding. This module also generates two types of features, \textit{i.e.}, the HR mask feature and the HR boundary feature. 
As illustrated in Figure~\ref{figure: network}, the high-resolution mask feature $X_{m}^{hr}$ and the high-resolution boundary feature $X_{b}^{hr}$ are defined as:
\begin{equation}
\left\{
\begin{aligned}
&X_{en}  = f_{m}^{en}(X_{e}^{1}) + B(X_{m}^{2}),\\
&X_{m}^{hr} =f_{m}^{hr}(X_{en}), \\
&X_{b}^{hr} =f_{b}^{hr}(X_{en}), 
\end{aligned}
\right.
\end{equation}
where $f_{m}^{en}$, $f_{m}^{hr}$, and $f_{b}^{hr}$ denote different mapping functions, respectively,  implemented by a convolutional layer. $B$ is a bilinear interpolation function.

\subsubsection{Camouflaged Instance Prediction}
To predict camouflaged instances, we design a multi-task learning module to aggregate the shared query and high-resolution mask features and boundary features for simultaneous camouflaged instance prediction and camouflaged instance boundary detection. 
As shown in Figure~\ref{figure: prediction}, we design three branches to predict the location confidence score of each instance, the binary mask, and the boundary detection. The first branch aims to predict the probability of the existence of an instance at each location ($L^{i}$), which is defined as follows: 
\begin{equation}
L^{i} = \texttt{MLP}_{\times 1}(Q_{mb}^{i}),
\end{equation}
The second branch is to aggregate high-resolution boundary features and learned query embeddings to predict the instance boundaries, calculated as: 
\begin{equation}
E^{i} = \texttt{MLP}_{\times 3}(Q_{mb}^{i}) \otimes X_{b}^{hr},
\end{equation}
where $\otimes$ denotes element-wise multiplication. The third branch is to aggregate high-resolution mask features and learned query embeddings to predict camouflaged instances, formulated as: 
\begin{equation}
M^{i} = \texttt{MLP}_{\times 3}(Q_{mb}^{i}) \otimes X_{m}^{hr},
\end{equation} 
The proposed module simultaneously predicts both camouflaged instances and camouflaged instance boundaries. The joint learning of these two tasks enables our model to better explore and interact with two types of cues, so as to learn a powerful representation of camouflaged instances and improve the performance.

\begin{table*}[t!]
\caption{Ablation experiments on the number of decoders, which also denotes the number of feature scales fed into the decoder.}
\centering
\renewcommand{\arraystretch}{1.0}
\setlength\tabcolsep{3.5mm}
\resizebox{1\textwidth}{!}{
\begin{tabular}{lccc|ccc|ccc}
\toprule
\rowcolor{gray!20}
 & &  & &\multicolumn{3}{c|}{COD10K} &\multicolumn{3}{c}{NC4K} \\ \cline{5-10}
 
\rowcolor{gray!20}
\multirow{-2}*{Models} & \multirow{-2}*{Features} &\multirow{-2}*{Parameters}
&\multirow{-2}*{FLOPs} & $\rm AP$ &$\rm AP_{50}$&$\rm AP_{75}$&  $\rm AP$ &$\rm AP_{50}$&$\rm AP_{75}$ \\
\midrule
Single-Scale (1)&$\{X_{4}\}$&32.5M&195.5G&44.4&71.0&45.6&47.2&73.7&49.5 \\
Multi-Scale (2)&$\{X_{4}, X_{3}\}$&36.0M&197.7G&43.9&70.7&44.8&47.1&73.9&49.6 \\
Multi-Scale (3)&$\{X_{4}, X_{3}, X_{2}\}$ &37.5M &221.0G &\textbf{45.2} &\textbf{71.6} &\textbf{46.6} &47.2 &74.2 &49.2 \\
Multi-Scale (4)&$\{X_{4}, X_{3}, X_{2}, X_{1}\}$ &40.0M &245.7G &44.3 &71.2 &46.0 &\textbf{47.5} &\textbf{74.8} &\textbf{49.9} \\
\bottomrule
\end{tabular}}
\label{table: decoder_num}
\end{table*}

\begin{table}[t!]
\small 
\centering
   \caption{Ablation experiments on the query initialization.}
   \renewcommand{\arraystretch}{1.0}
   {
\setlength\tabcolsep{7.8pt}
\begin{tabular}{cc|ccc|ccc}
    \toprule
    
    \rowcolor{gray!20}
&  &\multicolumn{3}{c|}{COD10K} &\multicolumn{3}{c}{NC4K} \\ \cline{3-8}
    
    \rowcolor{gray!20}
 \multirow{-2}*{$Q_{m}^{0}$}  & \multirow{-2}*{$Q_{b}^{0}$} &  $\rm AP$ &$\rm AP_{50}$&$\rm AP_{75}$&  $\rm AP$ &$\rm AP_{50}$&$\rm AP_{75}$ \\ 
 \midrule

B &B&44.6&71.2&45.6&46.2&72.8&48.1 \\
A&A&43.8&70.3&45.0&47.1&73.9&49.1 \\
\textbf{A }&\textbf{B }&\textbf{45.2}&\textbf{71.6}&\textbf{46.6}&\textbf{47.2}&\textbf{74.2}&\textbf{49.2} \\
B&A&44.1&70.8&45.4&46.8&73.4&48.4 \\
    \bottomrule
    \end{tabular}%
  \label{tab: queryinitialization}
  }
\end{table}%

\subsection{Loss Function}
We refer to the bipartite matching loss in DETR~\cite{carion2020detr} to match predictions and ground truths. For mask and boundary supervision, we adopt binary cross entropy (BCE) loss ($L_{bce}$) and Dice loss ($L_{dice}$). For location score supervision, we only use BCE loss ($L_{bce}^{loc}$). Thus, the total loss is defined as: 
\begin{equation}
\left\{
\begin{aligned}
& \mathcal{L}_{u}  = \lambda(\mathcal{L}_{bce} + \mathcal{L}_{dice}), u \in \{mask, boundary\} \\
& \mathcal{L}^{total} = \lambda_{loc} \mathcal{L}_{bce}^{loc} + \alpha \mathcal{L}_{mask} + \beta \mathcal{L}_{boundary}
\end{aligned},
\right.
\end{equation}
where $\lambda$, $\lambda_{loc}$, $\alpha$, and $\beta$ are the balancing factors. We set $\lambda$ and $\lambda_{loc}$ to 2 and 5 in our experiments.

\section{Experiments and Results}

\subsection{Implementation Details}

\subsubsection{Training Settings}
For a fair comparison, we implement our model using the widely used codebase Detectron2~\cite{wu2019detectron2}. In line with other competitors, we adopt ResNet-50 and ResNet-101~\cite{he2016deep} as the backbone. All backbones are initialized with weights trained on ImageNet-1k~\cite{krizhevsky2012imagenet}. We set the batch size to 16 and the number of training iterations to 15,000. The initial learning rate is 2.5e4. The resolutions are critical for training and testing. Following OSFomer~\cite{pei2022osformer}, we fixed the maximum size of the longest side to 1,333 and the minimum value of the short side to 400, 800. We also use scale jittering as data augmentation. 

\subsubsection{Datasets}
The existing instance-level COD datasets contain CAMO++ \cite{le2021camouflaged}, COD10K \cite{fan2020Camouflage}, and NC4K \cite{lv2021simultaneously}.
CAMO++ contains 5,500 image samples with instance-level annotations.
COD10K provides 3,040 images with instance-level annotations for training and 2,026 images for testing.
NC4K contains 4,121 instance-level annotation samples.
Since CAMO++ is not currently open source, we follow the settings in OSFormer and adopt the COD10K training set for model training, and then evaluate it on the COD10K and NC4K test sets.

\subsubsection{Evaluation Metrics}
We adopt the commonly used evaluation metrics in instance segmentation, including $\rm AP$, $\rm AP_{50}$, and $\rm AP_{75}$. Due to page limitations, more quantitative and qualitative results are provided in \textit{Supplementary Material}.

\subsection{Comparisons with State-of-the-Arts}
\subsubsection{Results on COD10K}
Table~\ref{table: all_data} shows the quantitative comparison of the proposed \ourmodel~and other competitors for camouflaged instance segmentation on COD10K. It can be seen that the proposed method consistently achieves the best performance on the three metrics. 
Specifically, for the backbone ResNet-50, in terms of $\rm AP$  and $\rm AP_{75}$, the proposed method achieves a performance gain of {10.2\% and 12.3\%}, respectively, compared to the general instance segmentation model - Mask2Former; and the performance gain of {9.2\% and 12\%}, respectively, compared to the camouflaged instance segmentation model - OSFormer.
For the backbone ResNet-101, compared with OSFormer, the proposed method improves the performance by {8.1\% and 11.9\%}, respectively, in terms of $\rm AP$ and $\rm AP_{75}$.
It is worth noting that the performance gain of $\rm AP_{50}$ is not significant compared to $\rm AP$ and $\rm AP_{75}$. This may be because most instances in these two datasets have IoU scores greater than 0.75. Overall, experiments demonstrate the performance superiority of the proposed method compared to other comparisons.

\subsubsection{Results on NC4K}
Table ~\ref{table: all_data} shows significant performance improvement of the proposed method on the NC4K dataset. Compared to OSFormer (ResNet-50), the performance gains are 11.5\%, 2.3\%, and 16.3\%, respectively, in terms of $\rm AP$, $\rm AP_{50}$ and $\rm AP_{75}$. Compared to the ResNet-101 version, the performance gains are 12.8\%, 4.2\%, and 17.1\%, respectively. 
When compared to Mask2Former, the performance gains are 5.8\%, 3.5\%, and 7.7\% respectively for the ResNet-50 backbone, and 1.8\%, 7.2\%, and 2.7\% respectively for the ResNet-101 backbone, in terms of $\rm AP$, $\rm AP_{50}$ and $\rm AP_{75}$.

\subsection{Ablation study}

\subsubsection{The Number of Scales in Decoder}
Multi-scale cross-attention is the key to updating query information, which can better aggregate the information of each instance. 
We test the performance of different scale settings on COD10K and NC4K datasets, shown in Table~\ref{table: decoder_num}. 
It can be seen that the proposed model achieves the best performance with 3 scales (``Multi-Scale (3)'') on the COD10K dataset, and with 4 scales (``Multi-Scale (4)'') on the NC4K dataset. However, increasing the number of feature scales inevitably leads to an increase in model complexity. We can see that although the addition of feature $X_1$ (``Multi-Scale (4)'') improves the performance on the NC4K dataset, the FLOPs also increase by 24.7G. Therefore, we choose the setting of ``Multi-Scale (3)'' in our experiments considering the trade-off of computational complexity and performance.

\begin{table*}[t!]
\centering
\caption{Comparison of different query update methods.}
\renewcommand{\arraystretch}{1.0}
\setlength\tabcolsep{5mm}
\resizebox{1\textwidth}{!}{
\begin{tabular}{lccc|ccc|ccc}
\toprule
\rowcolor{gray!20}
 & &  & &\multicolumn{3}{c|}{COD10K} &\multicolumn{3}{c}{NC4K} \\
 \cline{5-10}
 
\rowcolor{gray!20}
\multirow{-2}*{Models} & \multirow{-2}*{Backbone} &\multirow{-2}*{Parameters}
&\multirow{-2}*{FLOPs} & $\rm AP$ &$\rm AP_{50}$&$\rm AP_{75}$&  $\rm AP$ &$\rm AP_{50}$&$\rm AP_{75}$ \\
\midrule

Separation&ResNet-50&40.6M&206.8G&44.4&71.2&45.5&46.5&73.6&47.9 \\
Sharing&ResNet-50&35.7M&203.0G&44.1&70.8&45.3&\textbf{47.2}&\textbf{74.3}&49.1 \\
Ours&ResNet-50&37.5M&221.0G&\textbf{45.2}&\textbf{71.6}&\textbf{46.6}&\textbf{47.2}&74.2&\textbf{49.2} \\

\bottomrule
\end{tabular}}
\label{table: queries_update}
\end{table*}

\begin{table*}[t!]
\caption{Performance comparison under different backbones.}
\centering
\renewcommand{\arraystretch}{1.0}
\setlength\tabcolsep{5mm}
\resizebox{1\textwidth}{!}{
\begin{tabular}{lccc|ccc|ccc}
\toprule

\rowcolor{gray!20}
 & &  & &\multicolumn{3}{c|}{COD10K} &\multicolumn{3}{c}{NC4K} \\
 \cline{5-10}
 
\rowcolor{gray!20}
\multirow{-2}*{Models} & \multirow{-2}*{Backbone} &\multirow{-2}*{Parameters}
&\multirow{-2}*{FLOPs} & $\rm AP$ &$\rm AP_{50}$&$\rm AP_{75}$&  $\rm AP$ &$\rm AP_{50}$&$\rm AP_{75}$ \\
\midrule

\ourmodel&ResNet-50&37.5M&221.0G&45.2&71.6&46.6&47.2&74.2&49.2 \\
\ourmodel&ResNet-101&56.4M&272.1G&45.4&71.8&47.9&50.1&76.8&52.8 \\
\ourmodel&Swin-tiny&40.9M&212.9G&51.0&77.5&54.0&54.8&80.9&58.1 \\
\bottomrule
\end{tabular}}
\label{all_data}
\end{table*}

\begin{table}[t!]
\small 
  \centering
  \caption{Ablation study of $\alpha$ and $\beta$ in loss function.}
   \renewcommand{\arraystretch}{1.0}
   {
	\setlength\tabcolsep{7.8pt}
    \begin{tabular}{cc|ccc|ccc}
    \toprule
    
    \rowcolor{gray!20}
&  &\multicolumn{3}{c|}{COD10K} &\multicolumn{3}{c}{NC4K} \\
\cline{3-8}

    \rowcolor{gray!20}
 \multirow{-2}*{$\alpha$}  & \multirow{-2}*{$\beta$} &  $\rm AP$ &$\rm AP_{50}$&$\rm AP_{75}$&  $\rm AP$ &$\rm AP_{50}$&$\rm AP_{75}$ \\
\midrule

0.5&1&43.6&70.3&44.6&46.6&73.9&48.3 \\
1&1&44.7&71.5&45.8&46.7&73.2&48.9 \\
\textbf{1}&\textbf{2}&\textbf{45.2}&\textbf{71.6}&\textbf{46.6}&\textbf{47.2}&\textbf{74.2}&\textbf{49.2} \\
1&3&44.5&71.1&45.6&46.7&73.3&48.6 \\
    \bottomrule
    \end{tabular}%
  \label{tab: loss_w}}%
\end{table}%

\subsubsection{Query Initialization}
Effective queries can better explore instance semantics and accelerate model convergence. 
We test the performance of four kinds of initialization combinations, that is, random initialization~\cite{cheng2021per} for both mask and boundary queries (BB), our proposed salient points for both mask and boundary queries (AA), and the mixed initialization methods (AB and BA), shown in Table~\ref{tab: queryinitialization}. The best performance can be seen in the initialization combination of mask query using salient point selection and boundary query using random initialization. 
First, the same initialization method may lead to serious homogeneity, which reduces the generalization of the model. Second, the proposed model essentially achieves camouflaged instance segmentation and camouflaged boundary detection. These two tasks are interrelated, and the model is more inclined to camouflaged instance segmentation. Therefore, the points from the features with significant responses as the initialization of the mask queries can reduce the optimization difficulty, thereby improving the query representation. 
In another hand, the queries are derived from the highly responsive points in the features rather than directly resizing the feature map, which is more conducive to obtaining the location representation of the instance.

\subsubsection{Query Updating} 
The instance semantics explored by mask queries and boundary queries are different, and the unified update method is conducive to absorbing the advantages of both and obtaining a more powerful query representation. 
To verify the effectiveness, we adopt a separation updating strategy for comparison, denoted as ``separation''. Specifically, after cross-attention, the mask queries and boundary queries update independently without interaction.
As can be seen from Table~\ref{table: queries_update}, the effect of separation is significantly reduced. The lack of interaction between the two queries makes it difficult to model the relationship between instances, and the lack of boundary semantics makes background separation difficult.
Also, we compare another sharing queries method, denoted as ``sharing'', which only builds queries to jointly represent features of camouflaged instances and camouflaged instance boundaries. It shows good performance on the NC4K dataset. However, in essence, the sharing strategy is easy to confuse, with low generalization.
Therefore, there is no significant performance difference on the COD10K dataset. 
Overall, the proposed query update method achieves superior performance over other competitors. 

\begin{table}[t!]
\small 
  \centering
  \caption{Computational complexity comparison. All models are based on the ResNet-50 backbone. FLOPs are averaged under 100 samples. $\rm mAP$ is mean $\rm AP$  on COD10K and NC4K.}
  \renewcommand{\arraystretch}{1.0}
  {
	\setlength\tabcolsep{13.2pt}
    \begin{tabular}{c|ccc}
    \toprule
    
\rowcolor{gray!20}
Models &Parameters &FLOPs & $\rm mAP$ \\ \midrule
 
Mask R-CNN&43.9M&186.3G&26.4 \\

MS R-CNN&60.0M&198.5G&30.6\\

Cascade R-CNN&71.7M&334.1G& 27.4\\

HTC&76.9M&331.7G&29.0 \\

BlendMask&35.8M&233.8G&28.0\\

Mask Transfiner&44.3M&185.1G&29.1 \\
CondInst&34.1M&200.1G&32.0 \\
SOTR&63.1M&476.7G&28.6 \\
SOLOv2&46.2M&318.7G&33.5 \\
SparseInst&31.6M&165.8G&33.6 \\
Mask2Former&44.0M&230.0G&43.0 \\
\hline
OSFormer&46.6M&324.7G&41.8 \\
\ourmodel~(Ours)&37.5M&221.0G&\textbf{45.2} \\
\bottomrule
\end{tabular}%
\label{tab: flops}}%
\end{table}%

\subsubsection{Backbone}
As shown in Table~\ref{all_data}, ResNet-50 achieves an $\rm AP$ of 45.2\% on the COD10K dataset and reaches 47.2\% on the NC4K dataset. When increasing the convolution depth to 101, the $\rm AP$ is increased by 0.2\% on COD10K, and by 2.9\% on NC4K. Swin Transformer~\cite{liu2021swin}, as the current state-of-the-art Transformer model, further improves the camouflaged discrimination ability of the proposed model, with a performance improvement of 5.8\% on COD10K and a performance improvement of 7.6\% on NC4K.

\subsubsection{Hyperparameters in Loss Function}
To better train the model, we test different parameter settings in the loss function. As shown in Table~\ref{tab: loss_w}, our model achieves the best performance when $\alpha=1$ and $\beta=2$. Thus, we choose this parameter setting in our experiments.

\subsubsection{Computational Complexity}
Taking ResNet-50 as the backbone as an example, we report the comparison results of our model and various comparison models in terms of parameters and FLOPs in Table~\ref{tab: flops}, and also report the average $\rm AP$ results on the COD10K and NC4K datasets. Compared to OSFormer, the parameters are reduced by {19.5\%}, FLOPs are reduced by {31.9\%}, and $\rm AP$ is increased by {4.4\%}. By comparison, our model provides outstanding performance with relatively lightweight architecture. We can also see similar advantages compared to Mask2Former. 
The main reason is that our decoder is a lightweight architecture. The number of queries in the OSFormer greatly exceeds our model (2200 queries in OSFormer \textit{vs.} 20 queries in ours).

\section{Conclusion}
In this paper, we propose a novel unified query-based learning paradigm, called~{\ourmodel}, for camouflaged instance segmentation, which regards CIS as a query-based direct set prediction problem without other post-processing.
The proposed model integrates the two tasks of camouflaged instance segmentation and instance boundary detection using a multi-scale unified learning decoder and a multi-task learning module. Compared with the recently proposed OSFormer, our proposed UQFormer significantly improves the performance of camouflaged instance segmentation with a lower computational overhead. We hope that the proposed efficient model can serve as a new baseline for camouflaged instance segmentation and facilitate future research.

\bibliographystyle{ACM-Reference-Format}
\balance
\bibliography{sample-base}


\newpage

\section*{\huge Appendix}
\vspace{2mm}

In the appendix, we provide more details on our proposed \textit{UQFormer}, such as training setting, ablation studies, and qualitative comparison. 

\setcounter{section}{0}
\section{More Details}

\subsection{Training details}

During the training process, we follow the settings of Mask2Former and compute the loss function on the sampled K significant points, rather than on the entire prediction map. This will help reduce memory usage during training. In our experiments, instead of using the same set of points, we sample K point sets for the predicted masks and predicted boundaries, respectively, due to the differences in position for the camouflage instance masks and camouflaged boundary predictions. Specifically, we sample $112\times 112$ significant points according to the settings in Mask2Former.

\subsection{More details on salient points}
Inspired by PointRend, we first uniformly sample points on the feature map, and perform uncertainty sampling according to the saliency of each point. To ensure the diversity of sampling, some random points are added. We choose the L1 distance for saliency estimation. Then, we select the Topk salient points for mask query initialization, and aggregate the object instance features from the encoder via cross-attention, coupled with boundary query to model the relationship between object instances. Finally, location confidence and mask prediction are performed based on object queries. Our proposed method explores critical prior knowledge for object queries to improve query learning and accelerate model convergence. Besides, the mask query is dynamic and can be extended to a variety of scenarios, and can also reduce overfitting. Furthermore, boundary queries help to learn the semantic difference between objects and backgrounds for accurate segmentation.

\subsection{More details on closing module}
The closing operation first performs the \textit{padding} operation to prevent cross-border, then performs the \textit{erosion} operation (kernel size=3) to make the object edges clearer, and the \textit{dilation} operation (kernel size=3) to fill the hole inside the objects and remove the noises, and finally perform a \textit{padding} operation to remove added edge pixels.

\subsection{More details about query setting} The mask queries in the next stage are derived from the combined queries. The boundary queries of the next stage are derived from the boundary queries of the previous stage. In the ablation study, the ``separation" means that the boundary queries and mask queries interact independently with their respective features and then perform self-attention. In the ``sharing" setting, the mask and boundary queries share the same queries. In the prediction phase, the ``separation" is implemented as described.  In ``sharing",  mask and boundary queries (same queries) interact separately with the mask and boundary features and are supervised with their respective labels. The location score is directly predicted by the shared queries. The only difference between ``sharing" and ours is that the initialization of boundary queries for ``sharing" is a copy of mask queries. This may be the reason why the performance is similar to ours on NC4K. But our model significantly improves AP by 1.1\% on the COD10K. This verifies that the boundary queries mine more semantic cues for CIS.

\subsection{OSFormer vs. the proposed model} 
OSFormer only uses boundary labels as supervision for accurate segmentation. 
In our model, we introduce boundary cues into the object query to improve query feature representation, and combine mask queries and boundary queries for transformer decoding. Besides, a multi-task learning framework is used to make the two tasks mutually reinforcing.

\section{More Ablation Studies}

\subsection{Impact of query initialization on training}
\begin{figure}[b]
\centering
\small	
\begin{overpic}[width=1.\linewidth]{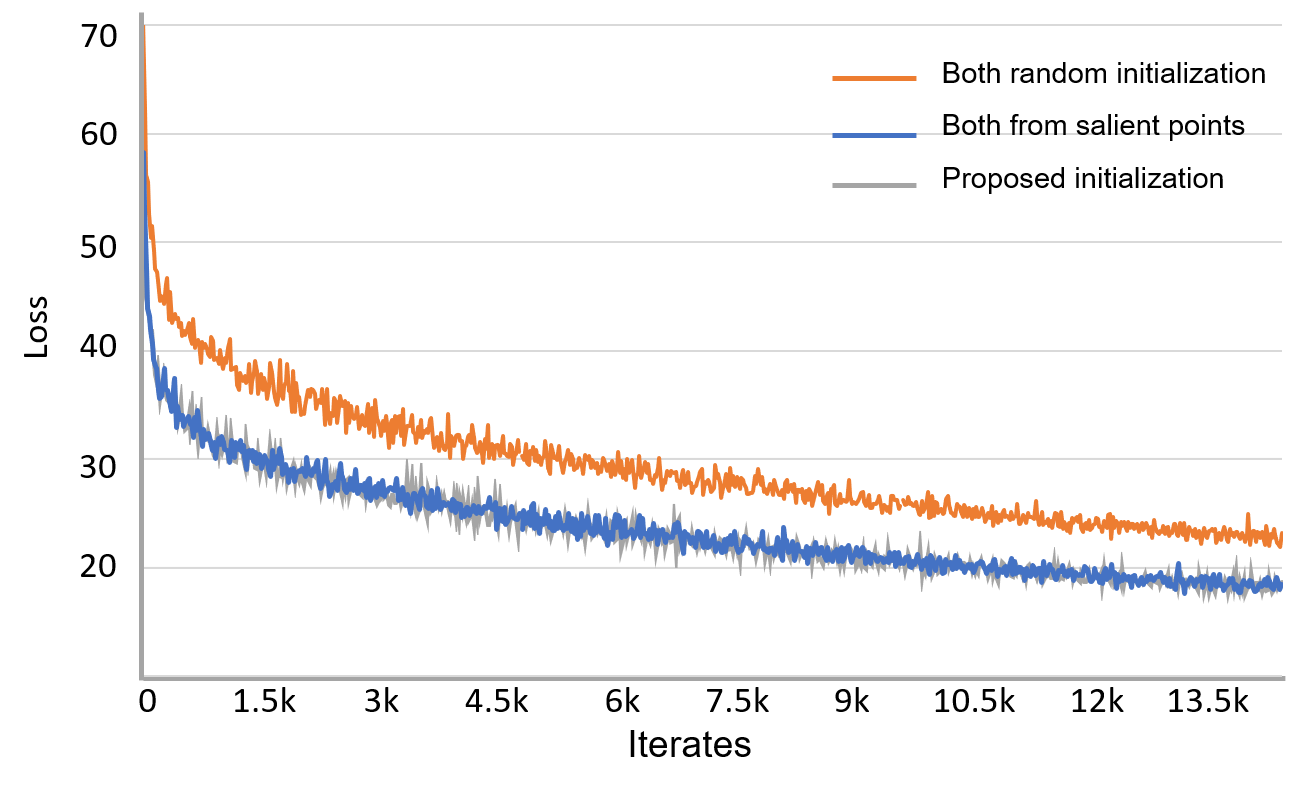}
\end{overpic}
\caption{The loss function curve of the proposed model under different initialization conditions.}
\label{figure:init}
\end{figure}

Figure~\ref{figure:init} shows the loss function curves of the proposed model under different initialization methods.
The query initialization method we propose not only removes the homogeneity between the mask query and the boundary query, but also provides a better convergence speed than the random initialization in Mask2Former. This is because the mask query comes from the sampling points with high confidence in the feature.

\begin{table*}[t!]
\centering
\caption{Ablation study on camouflaged instance boundary learning.}
\setlength\tabcolsep{4mm}
\renewcommand{\arraystretch}{1.2}
\resizebox{1\textwidth}{!}{
\begin{tabular}{lccc|ccc|ccc}
\hline
\multirow{2}*{Models} & \multirow{2}*{Backbone} &\multirow{2}*{Parameters}
&\multirow{2}*{FLOPs} &\multicolumn{3}{c|}{COD10K} &\multicolumn{3}{c}{NC4K} \\
& &  &  & $\rm AP$ &$\rm AP_{50}$&$\rm AP_{75}$&  $\rm AP$ &$\rm AP_{50}$&$\rm AP_{75}$ \\
\hline

w/o boundary learning &ResNet-50&37.4M&203.8G&44.0&71.0&44.8&45.7&72.0&47.6 \\

\ourmodel&ResNet-50&37.5M&221.0G&45.2&71.6&46.6&47.2&74.2&49.2 \\
\toprule
\end{tabular}}
\label{tab:bd}
\end{table*}

\subsection{Camouflaged instance boundary learning}

To verify the effectiveness of camouflaged instance boundary learning, we directly remove the camouflaged instance boundary detection branch.  As shown in Table~\ref{tab:bd}, AP drops by 1.2\% and AP$_{75}$ drops by 1.8\% on COD10K dataset. On the NC4K dataset, its AP decreased by 1.5\%, AP$_{50}$ decreased by 2.2\%, and AP$_{75}$ decreased by 1.6\%. 
This is because learning camouflaged instance boundaries facilitates capturing differential representations between the camouflaged objects and backgrounds, thereby better segmenting camouflaged objects from similar backgrounds. At the same time, camouflaged instance boundary detection can also effectively boost the separation of overlapping and close objects, avoiding the identification of multiple instances as one instance.

\section{More Visualizations}

\subsection{Multi-Scale Unified Learning Transformer Decoder} 
\begin{figure*}[t!]
\centering
    \small	\begin{overpic}[width=1.\linewidth]{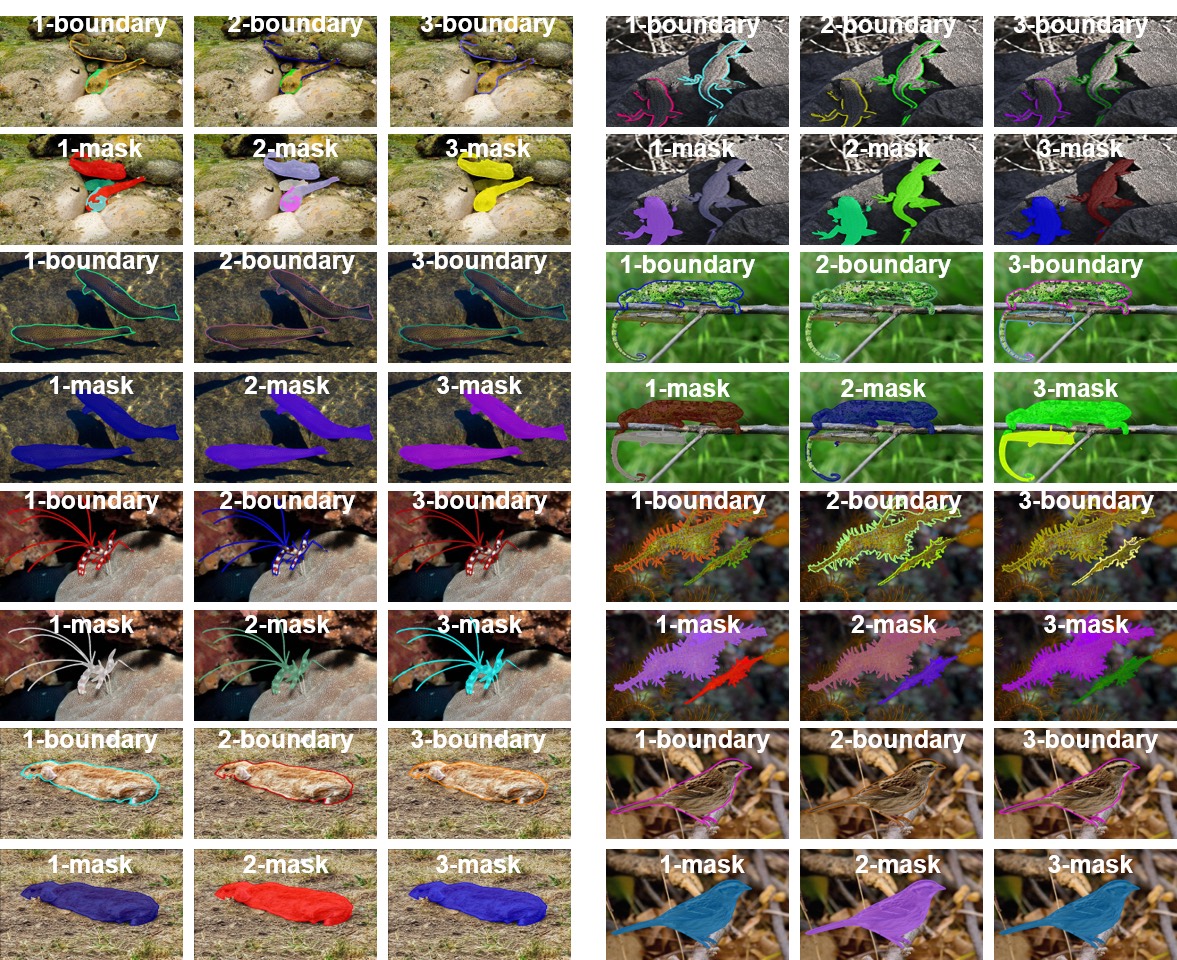}
    \end{overpic}
    \caption{Visualization of predictions in the multi-scale unified learning (MSUL) transformer decoder. We show the iterative optimization process of boundaries and masks at different scales. Among them, the text representation in each picture denotes  boundary prediction output by i-th scale and mask prediction output by i-th scale.}
    \label{fig:MSUL}
\end{figure*}

Figure~\ref{fig:MSUL} shows the predicted instance masks and instance boundaries after each iteration of multi-scale unified learning. It can be found that the model can basically capture the approximate information of the camouflaged instances after the first iteration. With more iterations, the model provides more accurate camouflaged instance boundary detection and mask prediction, and can also remove some false positive camouflage responses.

\subsection{Qualitative Comparison}
\begin{figure*}[t!]
	\centering
    \small	\begin{overpic}[width=1.\linewidth]{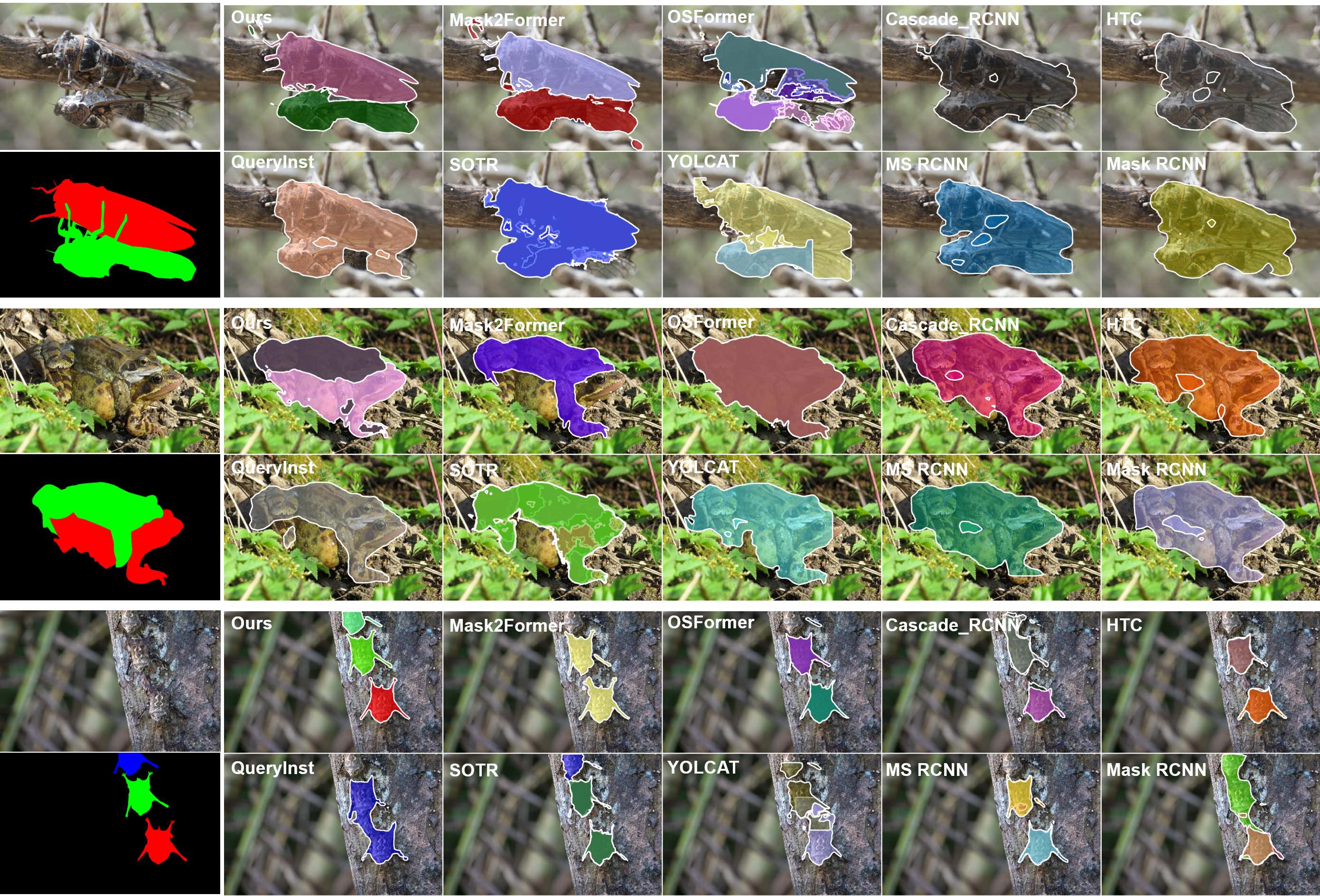}
    \end{overpic}
    \caption{Qualitative comparison of \ourmodel~and some representative methods.}
    \label{fig:visualcom}
    \vspace{-3mm}
\end{figure*}

We provide the qualitative comparisons on the COD10K and NC4K datasets, shown in Figure~\ref{fig:visualcom}. Obviously, our proposed method shows superior visual performance for more accurate camouflaged instance localization and segmentation.

\end{document}